%% file: main.tex
\documentclass[10pt,twocolumn,letterpaper]{article}

\usepackage[pagenumbers]{cvpr} 

\input{preamble}
\definecolor{cvprblue}{rgb}{0.21,0.49,0.74}
\usepackage[pagebackref,breaklinks,colorlinks,allcolors=cvprblue]{hyperref}

\def\paperID{} 
\def\confName{CVPR}
\def\confYear{2026}

\title{Dynamic Training-Free Fusion of Subject and Style LoRAs}

\author{
    Qinglong Cao$^{1,2}$ \hspace{4mm} Yuntian Chen$^{2}$\thanks{Corresponding author.} \hspace{4mm} Chao Ma$^{1}$ \hspace{4mm} Xiaokang Yang$^{1}$\\
    $^{1}$Shanghai Jiao Tong University \hspace{4mm} $^{2}$Eastern Institute of Technology, Ningbo \\
    \\
    {\footnotesize *Corresponding author}
    \vspace{-4.5mm}
}

\begin{document}

\maketitle
\input{sec/0_abstract}    
\input{sec/1_intro}
\input{sec/2_related_work}
\input{sec/3_method}
\input{sec/4_experiments}
\input{sec/5_conclusion}
{
    \small
    \bibliographystyle{ieeenat_fullname}
    \bibliography{main}
}

\input{sec/X_suppl}

\end{document}

%% file: sec/0_abstract.tex
\begin{abstract}
Recent studies have explored the combination of multiple LoRAs to simultaneously generate user-specified subjects and styles. However, most existing approaches fuse LoRA weights using static statistical heuristics that deviate from LoRA’s original purpose of learning adaptive feature adjustments and ignore the randomness of sampled inputs. To address this, we propose a dynamic training-free fusion framework that operates throughout the generation process. During the forward pass, at each LoRA-applied layer, we dynamically compute the KL divergence between the base model’s original features and those produced by subject and style LoRAs, respectively, and adaptively select the most appropriate weights for fusion. In the reverse denoising stage, we further refine the generation trajectory by dynamically applying gradient-based corrections derived from objective metrics such as CLIP and DINO scores, providing continuous semantic and stylistic guidance. By integrating these two complementary mechanisms—feature-level selection and metric-guided latent adjustment—across the entire diffusion timeline, our method dynamically achieves coherent subject-style synthesis without any retraining. Extensive experiments across diverse subject–style combinations demonstrate that our approach consistently outperforms state-of-the-art LoRA fusion methods both qualitatively and quantitatively.

\end{abstract}

%% file: sec/1_intro.tex
\section{Introduction}
\label{sec:intro}

 \begin{figure*}[h]
	\begin{center}
    \includegraphics[width=0.75\textwidth]{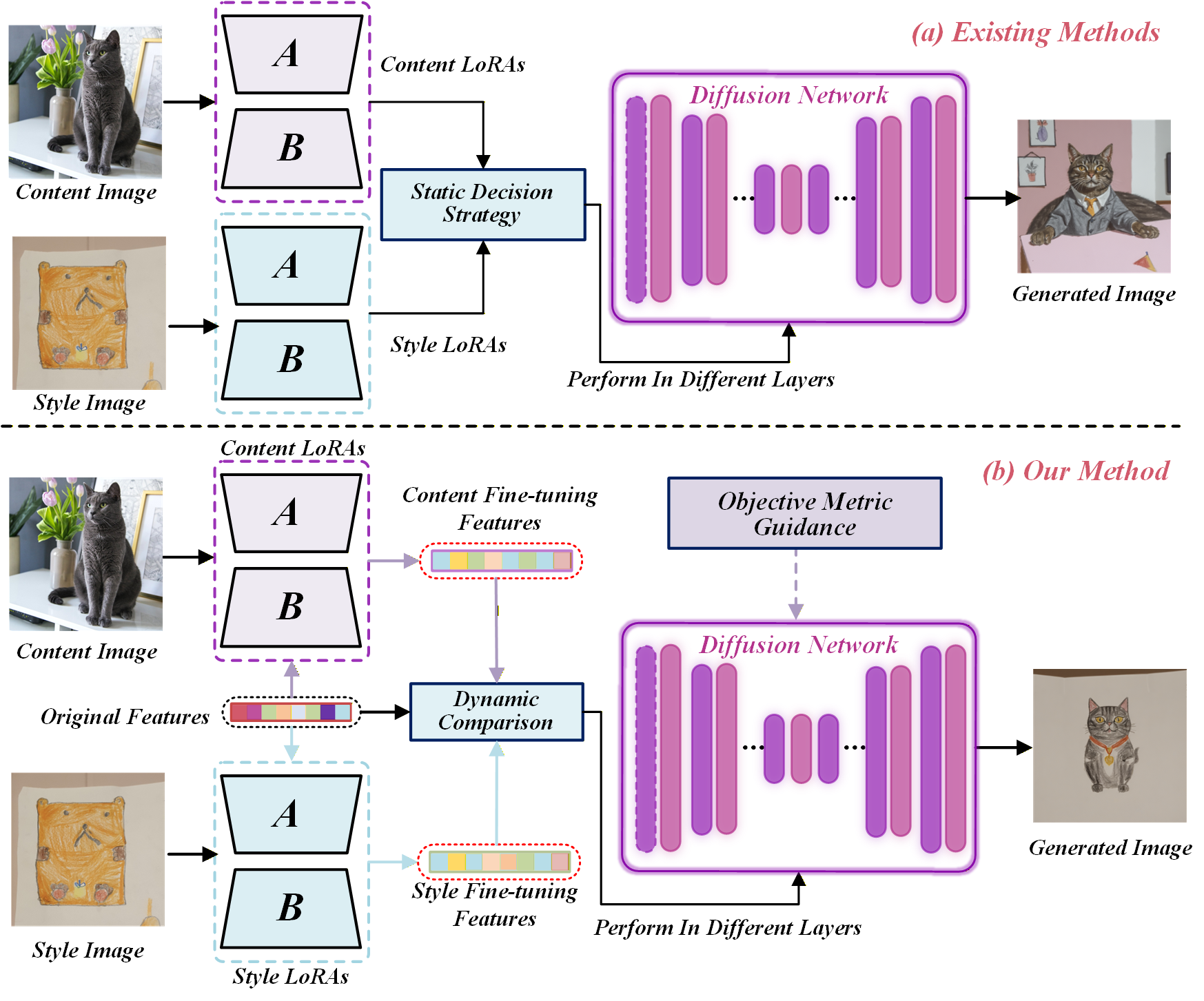}
	\end{center}
        \vspace{-6mm}
	\caption{(a) Existing methods directly rely on properties of LoRA weights to achieve fusion. (b) Our method integrates feature-level selection in the forward pass and latent-level refinement in the reverse process to enable dynamic training-free LoRAs fusion.}
	\label{fig1}
    \vspace{-5mm}
\end{figure*}

Diffusion models have demonstrated remarkable performance across a wide range of generative tasks~\cite{chen2024gentron,jiang2024scedit,gupta2024photorealistic,xing2024survey,zheng2023layoutdiffusion,ma2023diffusionseg,lu2025ovis2}. Among these, personalized image generation~\cite{ruiz2023dreambooth,sohn2023styledrop,xu2025personalized} has garnered increasing attention, as it requires the model to synthesize high-quality images that reflect user-specified content or style. Here, content refers to the semantic structure and subject identity, while style captures visual properties such as color, texture, and patterns. Although substantial progress has been made in generating images conditioned on either content or style alone, producing images that faithfully integrate both a specific subject and a specific style remains a challenging and unsolved problem.

Recently, Low-Rank Adaptation (LoRA)~\cite{hu2022lora} has emerged as a popular and versatile technique for parameter-efficient fine-tuning, making it particularly appealing for personalized generation tasks. Leveraging the modular nature of LoRA, recent studies have explored the fusion of independently fine-tuned LoRAs to jointly generate specific subjects in specific styles. For example, ZipLoRA~\cite{shah2024ziplora} proposes leveraging coefficient vectors to merge content and style LoRAs in each LoRA-applied layer. Differently, B-LoRA~\cite{frenkel2024implicit} investigates the impact of diverse LoRA layers and finds that modifying two distinct LoRA layers can effectively control the content and style of generated images. Furthermore, focusing on the intrinsic characteristics of LoRA weights, K-LoRA~\cite{ouyang2025k} selects LoRAs in each layer by comparing the Top-K elements of the weights. While these methods have demonstrated promising performance in LoRA fusion, as shown in Figure~\ref{fig1}~(a), their core strategies remain grounded in statistical properties of LoRA weights, which diverge from the original intent of LoRA—learning additional features to adapt to diverse functions. This divergence suggests that the fine-tuned features themselves, rather than the LoRA weights alone, are the true key to effective fusion. More importantly, previous static fusion methods ignore the randomness of sampled latent inputs during generation, limiting their adaptability and leading to suboptimal performance.

Inspired by these, as shown in Figure~\ref{fig1}~(b), we propose a dynamic training-free fusion framework that operates throughout the diffusion process. In contrast to prior work that treats LoRA fusion as a static weight blend, our approach reinterprets fusion as a dynamic representation-aware decision process aligned with the generative dynamics of diffusion models. In particular, the previous method~\cite{ouyang2025k} argues that the absolute values of LoRA weights indicate their importance in the diffusion process. By contrast, we propose that the feature changes induced by LoRAs serve as a more direct and key indicator of their impact. In each LoRA-applied layer during the forward pass, we compute the fine-tuned features from both the style LoRAs and the content LoRAs, respectively. To better quantify the extent of feature change, we leverage the  Kullback-Leibler~(KL) divergence between the fine-tuned features and the original features, determining which LoRA is more suitable for each layer based on the magnitude of distributional change. In this way, though the sampled inputs vary, the most representative content and style features are adaptively retained in each LoRA-applied layer.

Crucially, this layer-wise selection constitutes only the first half of our dynamic fusion strategy. To ensure global coherence and high-fidelity synthesis, we further introduce a metric-driven refinement mechanism that dynamically operates throughout the reverse denoising process. Specifically, we first generate two reference images using the subject and style LoRAs independently—serving as semantic and stylistic anchors. During denoising, at each timestep, we evaluate the intermediate prediction against these references using objective metrics such as CLIP~\cite{radford2021learning} and DINO~\cite{caron2021emerging} scores, which quantify content fidelity and style consistency, respectively. These objective metrics are then used to compute a composite guidance signal, whose gradient is applied to guide the latent trajectory toward regions that better align with the desired subject–style composition. This training-free, step-wise correction enables continuous refinement without any additional supervision.

By integrating feature-level selection in the forward pass and latent-level refinement in the reverse process, our method achieves effective integration of subject and style LoRAs, enabling high-quality image generation that preserves both subject fidelity and stylistic accuracy in a training-free manner. The main contributions of this work are summarized as follows:
\begin{itemize}
\item We introduce a dynamic training-free LoRA fusion framework that shifts the paradigm from static weight-level heuristics to input-adaptive, representation-aware decisions throughout the generation process.
\item A KL divergence-based strategy adaptively selects the most informative LoRA at each layer based on feature perturbation magnitude. Complementarily,  objective metric scores provide metric-guided refinement during denoising, enhancing both semantic and stylistic fidelity.
\item Our method is fully training-free, plug-and-play, and demonstrates superior performance across diverse subject–style combinations on multiple benchmarks, without requiring retraining or additional supervision.
\end{itemize}

%% file: sec/2_related_work.tex
\section{Related work}
\label{sec:related_work}
\textbf{Diffusion Models for Custom Generation.}
With the rapid development of diffusion models~\cite{he2024cameractrl,cao2024teaching,he2025cameractrl,lu2025ovis2, ho2022video,zhang2023uncovering,dong2025insight}, many researchers have introduced diverse approaches to fine-tuning large-scale diffusion models for custom generation, which aims to produce images of user-specified subjects or styles based on language descriptions. For instance, Textual Inversion~\cite{gal2022image} focuses on optimizing a single word embedding to capture unique and varied concepts. DreamBooth~\cite{ruiz2023dreambooth} designs text prompts containing a unique identifier to more effectively generate images with the desired subjects. CustomDiffusion~\cite{kumari2023multi} fine-tunes the cross-attention layers within the diffusion model to learn multiple concepts simultaneously. Additionally, some methods~\cite{avrahami2023break,shi2024instantbooth,xie2023smartbrush,xiao2024fastcomposer} achieve custom generation without additional training, yet these typically target specific single tasks. Recently, parameter-efficient fine-tuning techniques such as LoRA~\cite{hu2022lora} and StyleDrop~\cite{sohn2023styledrop} have gained popularity due to their ability to fine-tune models with low-rank adaptations, making them attractive for custom generation.

\textbf{LoRAs combination for image generation.}
Since the rise in popularity of LoRA applications~\cite{zhang2023lora,zhou2024lora,zi2023delta}, many studies on LoRA combinations have been proposed. Some methods~\cite{dong2024continually,gu2023mix,jiang2024mc,xing2024csgo} focus on fusing multiple object LoRAs, enabling diffusion models to generate various new concepts and replace these objects through masking strategies. Meanwhile, several advanced methods address content-style LoRA fusion. For instance, Mixture-of-Subspaces~\cite{wu2024mixture} designs learnable mixer weights to fuse various LoRAs; ZipLoRA~\cite{shah2024ziplora} leverages merge vectors across varying layers to linearly combine subject and style LoRAs; B-LoRA~\cite{frenkel2024implicit} investigates the impact of different LoRA layers and finds modifying two distinct layers can effectively control the content and style of generated images; and K-LoRA~\cite{ouyang2025k} selects the appropriate LoRAs in each layer by comparing top-K elements of different LoRA weights. Although these methods have shown promising performance, they directly rely on the properties of LoRA weights. The original intent of LoRA is to learn additional features to adapt to diverse tasks. Previous methods ignore the input randomness. Thus, dynamic fusion methods based on fine-tuned features could be more effective. To this end, we introduce a dynamic training-free LoRA fusion paradigm with input-adaptive, representation-aware decisions.

%% file: sec/3_method.tex
\section{Preliminaries}
\textbf{Diffusion Models.} Diffusion models~\cite{saharia2022palette, kazerouni2022diffusion, chen2023diffusiondet, amit2021segdiff} have demonstrated impressive performance across various generative tasks. They mainly consist of a forward noise addition process and a reverse denoising process. During the forward process, the original image is progressively transformed into Gaussian noise through incremental noise addition. In the reverse process, conditioned on paired text prompts, the diffusion network, typically a U-Net, gradually denoises the noisy input step-by-step, starting from randomly sampled pure noise. At inference time, the trained diffusion network achieves text-to-image generation based on the given textual input.

\textbf{LoRA.} Low-Rank Adaptation (LoRA~\cite{hu2022lora}) is a lightweight fine-tuning technique originally developed for large language and diffusion models. Rather than updating the full parameter matrix \(W_0\), LoRA exploits the observation that the update \(\Delta W\in\mathbb{R}^{m\times n}\) often lies in a low-dimensional subspace. Concretely, one factorizes \(\Delta W = BA\) with \(B\in\mathbb{R}^{m\times r}\) and \(A\in\mathbb{R}^{r\times n}\) for \(r\ll\min(m,n)\), and only \(A,B\) are learned while \(W_0\) remains fixed. The tuned model thus uses weights \(W_0 + BA\).
In our setting, let \(D\) be a diffusion model with base weights \(W_0\). To capture a new concept, we train a LoRA pair \(\Delta W_x\) so that the adapted model is \(D_x = W_0 + \Delta W_x\).

\textbf{Dynamic Representation Guidance.}  
The primary goal of LoRA is to learn fine-grained feature adjustments that guide a diffusion model toward specific behaviors. To leverage this capability, we first generate the modified features produced by both the content and style LoRAs. While prior work~\cite{ouyang2025k} uses the static magnitudes of LoRA weight updates as a proxy for their influence, we argue that the actual dynamic change in feature distributions provides a more direct and interpretable measure of impact since the randomly sampled inputs. We therefore quantify this change via Kullback–Leibler (KL) divergence between the original and LoRA-modified features, and use it to select the most informative branch for fusion at each layer.

In addition, objective metrics such as CLIP and DINO scores offer an effective way to assess the quality of LoRA fusion. Higher scores indicate better alignment with the intended semantics or style, and thus can serve as composite guidance signal to guide the diffusion network. The guidance score \(R\) based on the CLIP metric is computed as:
\begin{equation}
    R({\hat x_0}) = 1 - S_{\mathrm{CLIP}}(x_{\mathrm{ref}}, {\hat x_0}),
    \label{eq1}
\end{equation}
where \(S_{\mathrm{CLIP}}\) computes the CLIP similarity score, \(x_{\mathrm{ref}}\) is the reference image, and \({\hat x_0}\) is the predicted original image at step \(t\).

The guiding scores act as residuals that can be treated as virtually observed values~\cite{kaltenbach2020incorporating} with \({\hat R} = 0\) and virtual likelihood:
\begin{equation}
    p(\hat R = 0 \mid x_t) = \mathcal{N}(0 \mid R(x_0), \sigma_r^2 I),
    \label{eq2}
\end{equation}
where \(x_t\) is the intermediate result at timestep \(t\), and \(\sigma_r\) is a predefined constant controlling the enforcement strength of the virtual observation. Although residuals are minimized during the generation process, they are not guaranteed to reach zero. To incorporate this guidance into the diffusion process, we apply Bayesian rule:
\begin{equation}
    p(x_t \mid \hat R = 0) = \frac{p(x_t)\, p(\hat R = 0 \mid x_t)}{p(\hat R = 0)}.
\end{equation}
Taking the gradient of the log-likelihood with respect to \(x_t\), we obtain:
\begin{equation}
\begin{split}
    \nabla_{x_t} \log p(x_t \mid \hat{R} = 0)
    ={}& \nabla_{x_t} \log p(x_t) \\
    &+ \nabla_{x_t} \log p(\hat{R} = 0 \mid x_t),
\end{split}
\label{eq5}
\end{equation}

where the first term is the standard score function predicted by the diffusion model~\cite{song2020score}. For the second term, substituting Eq.~\ref{eq2} yields:
\begin{align}
    \nabla_{x_t} \log p(\hat{R} = 0 \mid x_t)
    &= -\frac{1}{\sigma_r^2} \nabla_{x_t} \lVert R(x_0) \rVert_2^2 \nonumber \\
    &\approx -\frac{1}{\sigma_r^2} \nabla_{x_t} \lVert R(\hat{x}_0) \rVert_2^2,
    \label{eq6}
\end{align}

where we approximate \(x_0\) with \({\hat x_0}\). In practice, because guidance scores range between \([0, 1]\), we simplify as:
\begin{equation}
    \nabla_{x_t} \log p(\hat R = 0 \mid x_t) \propto -\nabla_{x_t} R({\hat x_0}).
    \label{eq7}
\end{equation}
Thus, to implement the guidance, the denoising step is modified as:
\begin{equation}
    x_{t-1} = x_{t-1}^{\mathrm{ori}} - m \nabla_{x_t} R({\hat x_0}),
    \label{eq8}
\end{equation}
where \(x_{t-1}^{\mathrm{ori}}\) is the original output of the \(t\)-step denoising process, \({\hat x_0}\) is the predicted original image at step \(t\), and \(m\) is a predefined scaling factor.

 \begin{figure*}[h]
	\begin{center}
    \includegraphics[width=0.9\textwidth]{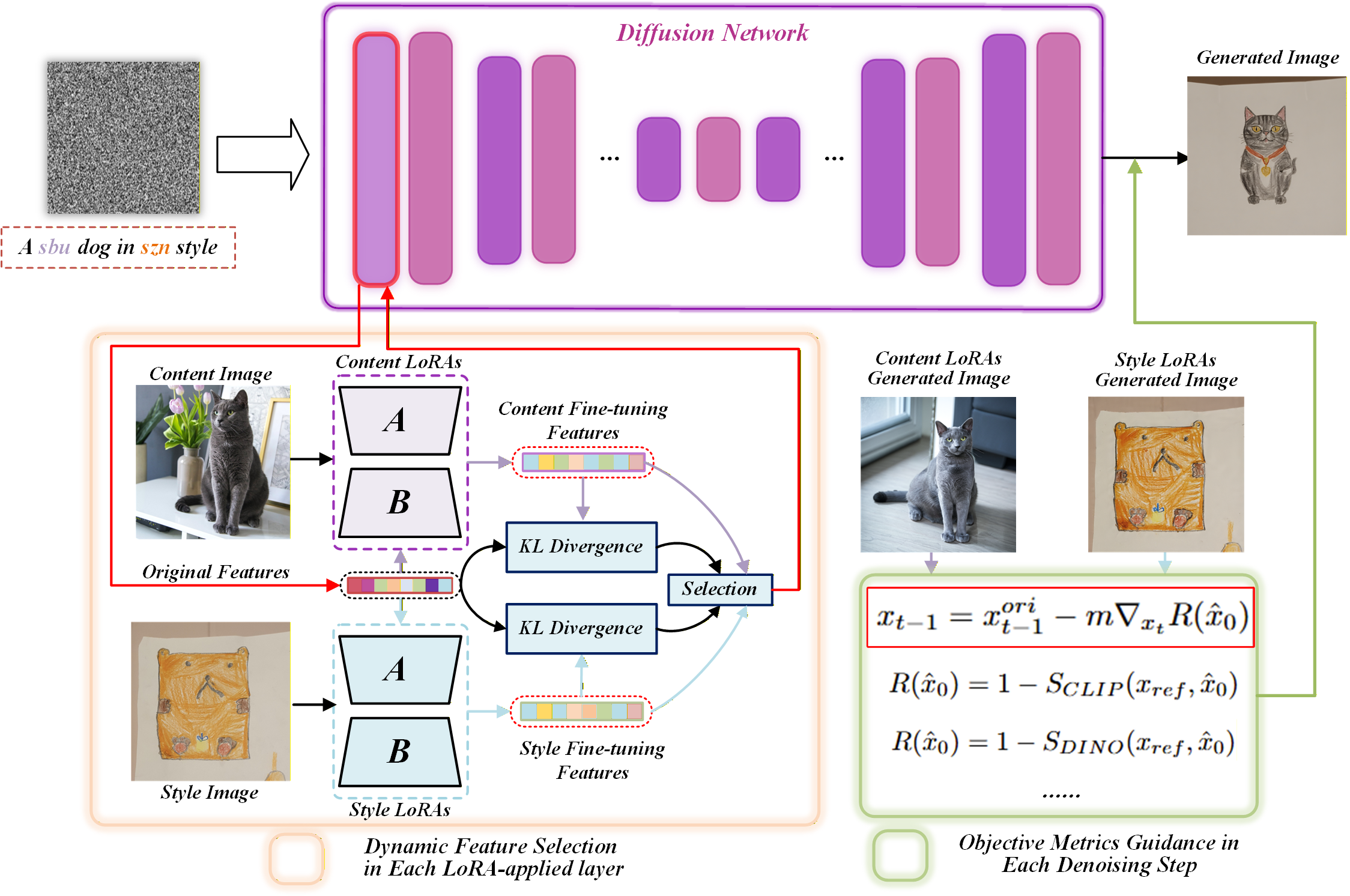}
	\end{center}
        \vspace{-3mm}
    \caption{Overview of our method. By performing dynamic feature selection based on representation perturbation and applying metric-guided refinement throughout the denoising process, our framework enables training-free fusion of subject and style LoRAs.}
	\label{fig2}
    \vspace{-3mm}
\end{figure*}

\section{Method}
As illustrated in Figure~\ref{fig2}, our framework guides a base diffusion model \(D\) to generate a specified subject in a specified style by dynamically fusing content and style LoRAs without any additional training. Let the base model \(D\) consist of pre-trained weights \(W_0^i\) at layer \(i\). Applying the LoRA \(L_x\) with weight updates \(\{\Delta W_x^i\}\) yields the adapted model:
\begin{equation}
  D_{L_x} = D \odot L_x = W_0 + \Delta W_x.  
\end{equation}
In the experimental setting, we are given the content LoRA weights \(L_c\{\Delta W_c^i\}\), the style LoRA weights \(L_s\{\Delta W_s^i\}\), and the base model \(D\). Rather than merging LoRA weights heuristically, we perform dynamic, representation-aware fusion throughout the diffusion process: at each LoRA-applied layer, we select the branch that induces more significant feature perturbation; during denoising, we refine the latent trajectory using objective metric scores. This two-stage, dynamic training-free strategy enables coherent integration of subject and style in a single forward pass.

\subsection{Feature-Level Selection During Forward Pass}
Prior work has used the absolute values of LoRA weight updates as a proxy for their importance in diffusion models~\cite{ouyang2025k}. However, the core function of LoRA is to induce feature-level adjustments rather than merely altering weight magnitudes, and the static methods ignore the randomness of sampled inputs. Motivated by this principle, we explicitly examine how content and style LoRA weight updates, \(\Delta W_c^i\) and \(\Delta W_s^i\), affect the base network’s \(i\)-th layer. Specifically, we apply these updates to the base weights \(W_0^i\) and compute the corresponding fine-tuned feature maps:
\begin{align}
  \hat F_c^{\,i+1} &= \bigl(W_0^i + \Delta W_c^i\bigr)\,F_i,\\
  \hat F_s^{\,i+1} &= \bigl(W_0^i + \Delta W_s^i\bigr)\,F_i,
\end{align}
where \(F_i\) represents the original features at layer \(i\).

To quantify the impact of these modifications, we compute the KL divergence between each fine-tuned feature distribution and the original feature distribution \(F_{i+1}\):
\begin{align}
  d_c^i &= \mathrm{KL}\bigl(\hat F_c^{\,i+1} \,\|\, F_{i+1}\bigr),\\
  d_s^i &= \mathrm{KL}\bigl(\hat F_s^{\,i+1} \,\|\, F_{i+1}\bigr).
\end{align}
We then compare \(d_c^i\) and \(d_s^i\) to determine which adjustment induces a more significant feature change:
\begin{equation}
  F^{\,i+1} \;=\;
  \begin{cases}
    \hat F_c^{\,i+1}, & \text{if } d_c^i \ge d_s^i,\\
    \hat F_s^{\,i+1}, & \text{otherwise}.
  \end{cases}
\end{equation}
By performing this selection at each layer, we retain the most impactful content or style information, enabling a dynamic, training-free fusion of subject and style LoRAs.

In contrast to weight-based fusion strategies, which are static and input-agnostic, our feature-based approach dynamically adapts to the input: as the prompt changes, so do the feature distributions and thus the fusion decisions. This input-conditional mechanism allows our method to flexibly handle generation tasks with diverse and evolving requirements.

\subsection{Latent-Level Refinement In Denoising Stage}
As discussed in the preliminaries, objective metrics such as CLIP and DINO scores effectively assess the quality of LoRA fusion, where higher scores indicate better alignment with the desired content or style. To obtain the guidance scores, we first leverage the content LoRAs \(\Delta W_c\) and style LoRAs \(\Delta W_s\) with content descriptions \(l_c\) and style descriptions \(l_s\) to generate the new reference content image \(I_c^{\mathrm{ref}}\) and reference style image \(I_s^{\mathrm{ref}}\):
\begin{equation}
    I_c^{\mathrm{ref}} = (W_0 + \Delta W_c)[l_c]
\end{equation}
\begin{equation}
    I_s^{\mathrm{ref}} = (W_0 + \Delta W_s)[l_s]
\end{equation}
With the original output \(x_{t-1}^{\mathrm{ori}}\) at the \(t\)-th denoising step and the predicted original image \(\hat{x}_0\) at step \(t\), we compute the CLIP and DINO scores~\cite{shah2024ziplora} to evaluate the prediction performance. The scores are obtained by extracting embeddings from the generated and reference images and computing cosine similarity:
\begin{equation}
    S_{\mathrm{CLIP}}^{\mathrm{content}} = \mathrm{Sim}_{\mathrm{cos}}\big(E_{\mathrm{CLIP}}(I_c^{\mathrm{ref}}), E_{\mathrm{CLIP}}(\hat{x}_0)\big)
\end{equation}
\begin{equation}
    S_{\mathrm{CLIP}}^{\mathrm{style}} = \mathrm{Sim}_{\mathrm{cos}}\big(E_{\mathrm{CLIP}}(I_s^{\mathrm{ref}}), E_{\mathrm{CLIP}}(\hat{x}_0)\big)
\end{equation}
\begin{equation}
    S_{\mathrm{DINO}}^{\mathrm{style}} = \mathrm{Sim}_{\mathrm{cos}}\big(E_{\mathrm{DINO}}(I_s^{\mathrm{ref}}), E_{\mathrm{DINO}}(\hat{x}_0)\big)
\end{equation}
where \(E_{\mathrm{CLIP}}\) and \(E_{\mathrm{DINO}}\) denote the image encoders of CLIP~\cite{radford2021learning} and DINO~\cite{caron2021emerging}, respectively. To compute the final guidance score, we evenly weight the three metrics:
\begin{equation}
    R(\hat{x}_0) = 1 - \frac{S_{\mathrm{CLIP}}^{\mathrm{content}} + S_{\mathrm{CLIP}}^{\mathrm{style}} + S_{\mathrm{DINO}}^{\mathrm{style}}}{3}
\end{equation}
With guidance scores and empirically setting scaling factor \(m = 10\), we guide the diffusion step as:
\begin{equation}
    x_{t-1} = x_{t-1}^{\mathrm{ori}} - m\nabla_{x_t} R(\hat{x}_0)
    \label{eq8}
\end{equation}
This training-free refinement is dynamically applied at every denoising step, providing continuous, objective feedback that steers the generation toward both subject fidelity and stylistic accuracy—without any retraining or additional supervision.

In summary, our method achieves dynamic training-free subject–style LoRA fusion by operating throughout the diffusion process: it performs dynamic feature selection in the forward pass based on KL divergence to retain the most informative representation at each layer, and applies metric-guided latent refinement during denoising using objective metrics feedback to ensure global semantic and stylistic coherence. This dual-stage, input-adaptive strategy requires no retraining, yet enables high-fidelity, plug-and-play composition of independently trained LoRAs.

%% file: sec/4_experiments.tex
\begin{table}[t!]
    \centering
    \footnotesize
        \begin{tabular}{lccc}
            \toprule
            \textbf{Method} & \textbf{Style Sim $\uparrow$} & \textbf{CLIP Score $\uparrow$} & \textbf{DINO Score $\uparrow$} \\
            \midrule
            \textbf{Direct} & 48.9\% & 66.6\% & 43.0\% \\
            \textbf{B-LoRA}~\cite{frenkel2024implicit} & 58.0\% & 63.8\% & 30.6\% \\
            \textbf{ZipLoRA}~\cite{shah2024ziplora} & 60.4\% & 64.4\% & 35.7\% \\
            \textbf{K-LoRA}~\cite{ouyang2025k} & 58.7\% & 69.4\% & 46.9\% \\
            \textbf{Ours} & 63.0\% & 78.5\% & 43.3\% \\
            \bottomrule
        \end{tabular}%
    \caption{Comparison of alignment results. Direct denotes direct arithmetic merging. }
    \label{table:alignment}
\end{table}

\section{Experiments}

\begin{table*}[t!]
\centering
\small
\begin{tabular}{lccc}
\toprule
\textbf{Method} & \textbf{User Preference} & \textbf{GPT-4o Feedback} & \textbf{Qwen2.5-VL Feedback} \\
\midrule
\textbf{ZipLoRA}~\cite{shah2024ziplora}          &       13.80\%      &       20.13\%   &   3.40\%  \\
\textbf{B-LoRA}~\cite{frenkel2024implicit}           &      21.89\%       &     11.67\%  &    9.11\%    \\
\textbf{K-LoRA}~\cite{ouyang2025k}           &       11.11\%        &     12.56\%  &   21.82\%     \\
\textbf{Ours}             &      53.20\%          &     55.64\%    &     65.67\%   \\
\bottomrule 
\end{tabular}
\caption{The Performance Comparison of user study results, GPT-4o and Qwen2.5-VL feedback.}
\label{tab:user_study}
\end{table*}

\begin{figure*}[t!]
    \centering
    \includegraphics[width=0.95\linewidth]{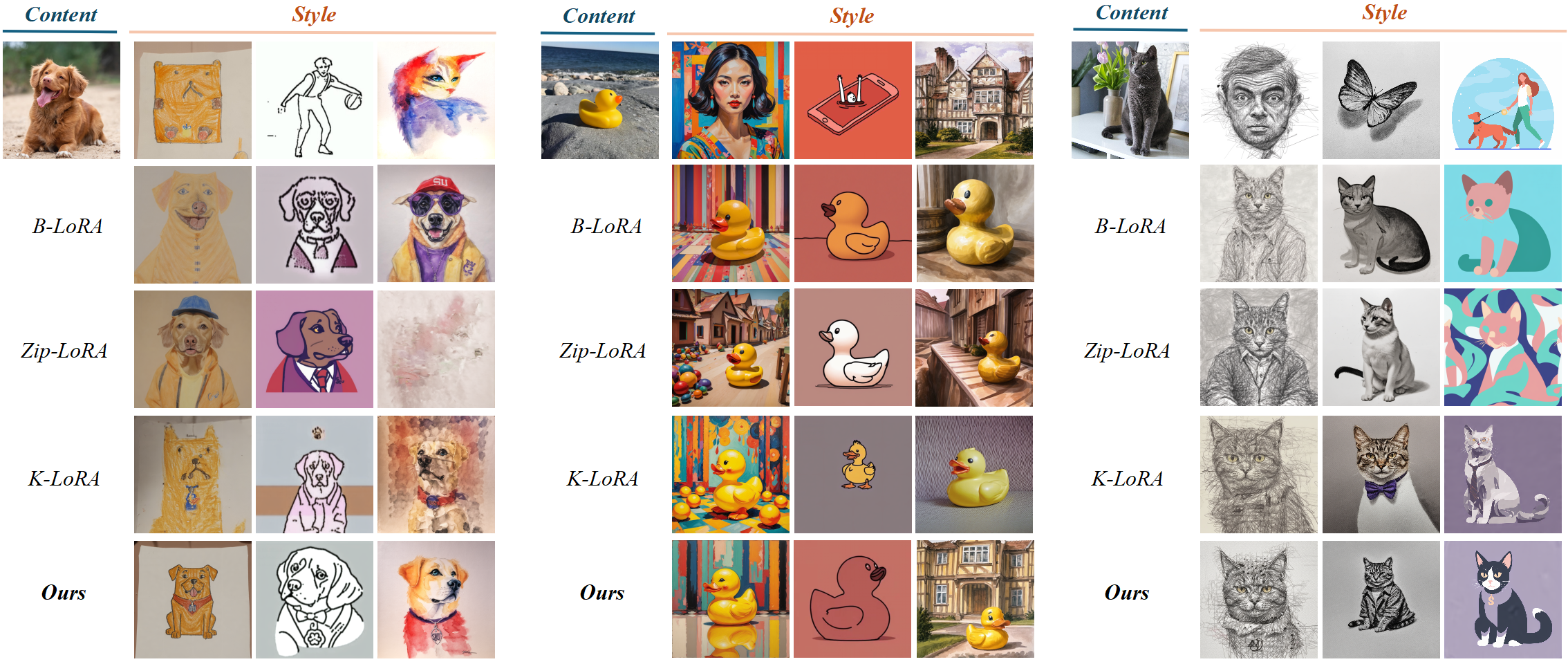}
    \caption{\textbf{Qualitative comparisons.}~We present images generated by our method and the compared advanced generation methods. Through nput-adaptive, representation-aware decisions throughout the generation process, our method effectively enables training-free fusion of subject and style LoRAs.
    }
    \label{fig:compare}
\end{figure*}

We evaluate the proposed LoRAs fusion approach under the experimental setup established by previous methods including K-LoRA, ZipLoRA, B-LoRA. Specifically, our method is applied to both the Stable Diffusion XL v1.0 base model and the FLUX model.

\textbf{Datasets.} For training the local LoRAs, we follow the convention of previous works~\cite{ouyang2025k, shah2024ziplora}. To train content LoRAs, we select diverse image sets from the DreamBooth dataset~\cite{ruiz2023dreambooth}, where each instance is represented by 4–5 images. For style LoRAs, we adopt the dataset introduced by the StyleDrop authors~\cite{sohn2023styledrop}, which includes a wide variety of stylistic exemplars spanning classical art to modern creative styles. Each style LoRA is trained using a single reference image.

\textbf{Implementation Details.}To obtain local LoRAs for experiments, we adopt the K-LoRA~\cite{ouyang2025k} strategy to fine-tune the SDXL v1.0 base model using a low-rank adaptation with rank set to 64. The LoRA weights—both style and content—are optimized using the Adam optimizer over 1000 steps with a batch size of 1 and a learning rate of 5e-5. For the FLUX model, we utilize publicly available, well-trained community LoRA weights obtained from HuggingFace. The corresponding experimental results on FLUX are provided in the supplementary material.

\subsection{Results}

\subsubsection{Quantitative Comparisons}
To objectively evaluate the performance of our training-free generation method, we adopt commonly used metrics from prior works, including Style Similarity, CLIP Score, and DINO Score, to assess the quality of the generated images. Following previous methods, we randomly selected 30 unique content–style pairs, each of which consists of 10 images to perform quantitative comparisons. Specifically, CLIP~\cite{radford2021learning} is employed to evaluate both the style alignment (\textit{Style Sim}) and content preservation (\textit{CLIP Score}), while DINO~\cite{zhang2022dino} is used to measure content consistency via \textit{DINO Score}. 
Table~\ref{table:alignment} presents a detailed comparison between our method and existing state-of-the-art methods. Our method achieves the best performance in both \textit{Style Sim} (63.0\%) and \textit{CLIP Score} (78.5\%). Notably, it provides a substantial improvement of 9.1\% in \textit{CLIP Score} compared to the strongest baseline. Although our method does not achieve the top performance in \textit{DINO Score}, it still ranks second, demonstrating a strong overall balance between style and content fidelity.
These results validate the effectiveness of our dynamic fusion method. As a supplementary verification, we report further evaluations with extra metrics in the supplmentary material, which also demonstrate the effectiveness of our proposed method.

\begin{figure*}[t!]
    \centering
    \includegraphics[width=1.0\linewidth]{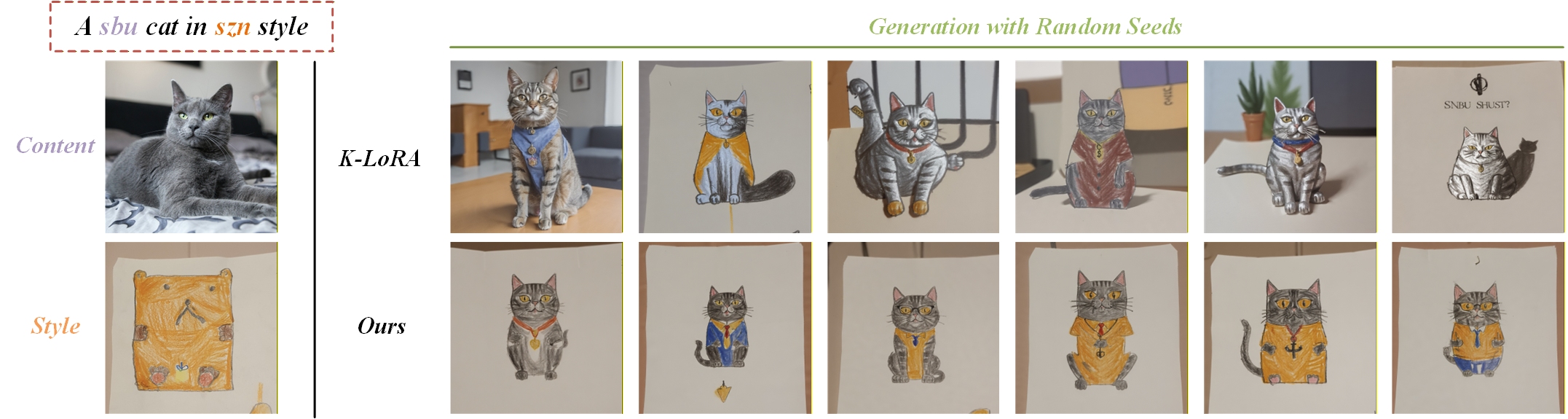}
    \caption{\textbf{Robustness Analysis.}~We present images generated by our method and K-LoRA with random seeds to analyze the robustness.
    }
    \label{fig:robu}
\end{figure*}

\subsubsection{User Study and MLLM-based Evaluations}
To further assess the perceptual quality of generated images beyond conventional metrics, we conduct a comprehensive user study and multimodal large language model (MLLM)-based evaluations. As shown in Table~\ref{tab:user_study}, we collect human preferences and automatic feedback from two strong MLLMs, GPT-4o~\cite{openai2024gpt4o} and Qwen2.5-VL~\cite{Qwen2.5-VL}. In the user study, participants were asked to choose their preferred images from outputs of four competing methods. Our proposed method is overwhelmingly favored, receiving 53.20\% of total votes, outperforming all baselines. Similarly, in LLM-based evaluations, our proposed method is consistently ranked highest, achieving 55.64\% preference by GPT-4o and an even more substantial 65.67\% by Qwen2.5-VL. These results not only confirm the good quality of generated images in terms of human preference but also highlight our method's effectiveness in producing stylistically and semantically coherent outputs that align well with multi-modal models. More setup details are provided in the supplementary material.

\begin{table}[t!]
    \centering
    \footnotesize
        \begin{tabular}{lccc}
            \toprule
            \textbf{Method} & \textbf{Style Sim $\uparrow$} & \textbf{CLIP Score $\uparrow$} & \textbf{DINO Score $\uparrow$} \\
            \midrule
             \textbf{Baseline} & 60.3\% & 75.6\% & 40.1\% \\
            \textbf{Only with FLS} & 59.5\% & 75.9\% & 43.7\% \\
            \textbf{Only with LLR} & 66.1\% & 78.9\% & 40.7\% \\
            \textbf{Ours(FLS+LLR)} & 64.0\% & 79.1\% & 43.4\% \\
            \bottomrule
        \end{tabular}%
    \caption{The ablation study of different components, including Feature-Level Selection (FLS) and Latent-Level Refinement (LLR). }
    \label{table:knowledge}
    \vspace{-3mm}
\end{table}

\subsubsection{Qualitative comparisons}
To visually assess the performance of different LoRA fusion methods, we present qualitative comparisons in Figure~\ref{fig:compare}. Overall, our training-free LoRA fusion method demonstrates superior visual quality, effectively preserving both content and style information. In contrast, most existing methods tend to retain content reasonably well but struggle to capture the target style faithfully. For example, in the second row and third column, B-LoRA correctly identifies the `dog' content but incorrectly applies a pink color inconsistent with the reference style. A similar issue is observed with K-LoRA in the fourth row and third column. Moreover, in the second-to-last row, K-LoRA produces a `cat' whose head and body exhibit inconsistent styles, indicating a failure in achieving global style coherence. In the eighth row, K-LoRA also fails to preserve the oil painting style entirely. Interestingly, Zip-LoRA, despite being the only method with learnable parameters, performs relatively worse. It often fails to capture the desired style (e.g., the `dog' in the fourth column) or generates semantically inaccurate content. These qualitative results further support the effectiveness of our dynamic fusion strategy throughout the generation process, which consistently delivers visually coherent outputs without the need for additional training.

\begin{table}[t!]
    \centering
    \footnotesize
        \begin{tabular}{lccc}
            \toprule
            \textbf{Divergence} & \textbf{Style Sim $\uparrow$} & \textbf{CLIP Score $\uparrow$} & \textbf{DINO Score $\uparrow$} \\
            \midrule
            \textbf{KL} & 59.5\% & 75.9\% & 43.7\% \\
            \textbf{JS} & 59.3\% & 75.3\% & 43.1\% \\
            \textbf{Cosine Similarity} & 58.9\% & 75.4\% & 43.3\% \\
            \textbf{Dot Product} & 58.4\% & 74.9\% & 43.8\% \\
            \bottomrule
        \end{tabular}%
    \caption{The ablation study of selection criteria in feature-level selection, including Kullback-Leibler (KL), Jensen-Shannon (JS) divergence, cosine similarity, and dot product. }
    \label{table:kl_vs_JS}
\end{table}

\begin{table}[t!]
    \centering
    \footnotesize
        \begin{tabular}{lccc}
            \toprule
            \textbf{Scaling Factor \(m\)} & \textbf{Style Sim $\uparrow$} & \textbf{CLIP Score $\uparrow$} & \textbf{DINO Score $\uparrow$} \\
            \midrule
            \textbf{1} & 62.1\% & 72.8\% & 32.4\% \\
            \textbf{5} & 64.8\% & 74.1\% & 32.3\% \\
            \textbf{10} & 64.0\% & 79.1\% & 43.4\% \\
            \textbf{20} & 66.8\% & 76.8\% & 35.1\% \\
            \bottomrule
        \end{tabular}%
    \caption{The ablation study of scaling factor \(m\) for Latent-Level Refinement. }
    \label{table:scaling factor}
    \vspace{-4mm}
\end{table}

\begin{figure*}[t!]
    \centering
    \includegraphics[width=1.0\linewidth]{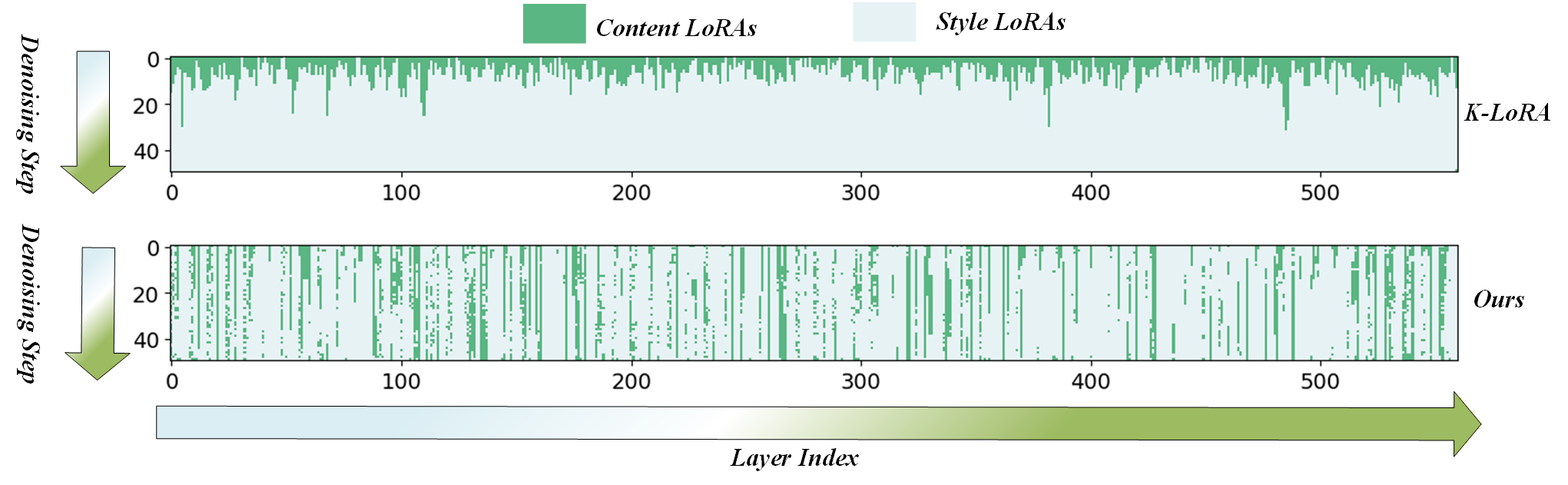}
    \caption{\textbf{LoRA Selection During the Generation Process.} The vertical axis represents 50 diffusion steps, while the horizontal axis denotes varying LoRA layers. Dark green indicates the selection of subject LoRA, and light blue indicates the selection of style LoRA.}
    \label{fig:lora_layers}
\end{figure*}

\subsection{Ablation Studies}

To thoroughly analyze the contributions of each component in our dynamic training-free fusion framework, we conduct extensive ablation studies across three aspects: (1) the effect of feature-level selection and latent-level refinement, (2) the choice of feature divergence metric for selection, and (3) the impact of the guidance scaling factor. For each setting, we randomly sample 25 subject–style combinations to ensure robustness.

\textbf{Effect of Feature-Level Selection and Latent-Level Refinement.}
Table~\ref{table:knowledge} presents the performance of different configurations. Using only feature-level selection (based on KL divergence) improves the \textit{DINO Score} by 3.6\% over the baseline, indicating better content consistency. In contrast, latent-level refinement alone significantly boosts \textit{Style Similarity} (66.1\%) and \textit{CLIP Score} (78.9\%), reflecting stronger semantic and stylistic alignment. When both components are combined, our method achieves the best overall performance—particularly in \textit{CLIP Score} (79.1\%) and \textit{DINO Score} (43.4\%)—demonstrating their complementary roles in local feature adaptation and global generation guidance.

\textbf{Choice of Selection Criteria.}
We evaluate alternative criteria for feature-level selection, including Kullback–Leibler (KL) divergence, Jensen–Shannon (JS) divergence, cosine similarity, and dot product (Table~\ref{table:kl_vs_JS}). Both KL and JS divergence yield superior performance compared to similarity-based metrics. KL divergence slightly outperforms JS in \textit{DINO Score} (43.7\% vs. 43.1\%), likely because it is more sensitive to asymmetric distribution shifts—making it better suited for detecting meaningful feature perturbations in content representation.

\textbf{Effect of Scaling Factor.}
As shown in Table~\ref{table:scaling factor}, we also explore the sensitivity of the scaling factor that balances the influence of latent-level refinement. Setting this factor too low (e.g., 1) leads to suboptimal \textit{CLIP} and \textit{DINO Scores}, while extremely high scaling (e.g., 20) causes instability in \textit{DINO Score}. A moderate value of 10 yields the best trade-off, with a strong boost in both \textit{CLIP Score} (79.1\%) and \textit{DINO Score} (43.4\%), highlighting the importance of proper guidance calibration.

\textbf{Robustness Analysis.} To further evaluate the robustness of our proposed framework, we visualize generated results under different random seeds and compare them with K-LoRA in Figure~\ref{fig:robu}. While K-LoRA exhibits significant variability in scene layouts and fails to consistently preserve the target style, our method maintains both the semantic content and stylistic attributes across different sampling conditions. This stability under stochastic perturbations highlights the effectiveness of input-adaptive, representation-aware modulation in enforcing coherent and reliable image generation.

\textbf{Visualization of Dynamic Feature Selection.}
Figure~\ref{fig:lora_layers} illustrates the dynamic feature-level selection mechanism employed by our method. Unlike K-LoRA, which relies solely on static weight elements for selection, our method performs input-conditional selection, dynamically choosing the more relevant LoRA weights based on input-dependent features. This adaptive strategy enables better alignment with the input semantics, thereby facilitating more effective and coherent style-content fusion. The superior performance observed in previous quantitative and qualitative comparisons further validates the advantages of this training-free, input-adaptive selection scheme.

%% file: sec/5_conclusion.tex
\section{Conclusion}

In this paper, we propose a dynamic training-free method for fusing subject and style LoRAs through put-adaptive, representation-aware decisions across the generation process. Specifically, during the forward pass, rather than relying on static weight-level heuristics, our approach adaptively selects LoRA branches at each layer based on the magnitude of feature perturbation—quantified via KL divergence. Moreover, in the denoising stage, objective metrics such as the CLIP and DINO scores are leveraged to dynamically refine the denoising trajectory using gradient-based feedback. This dual-stage strategy enables coherent integration of subject identity and artistic style without any retraining or additional supervision. Extensive experiments across multiple benchmarks demonstrate that our method consistently outperforms existing LoRA fusion approaches in both qualitative realism and quantitative metrics.



%% file: sec/X_suppl.tex
\clearpage
\setcounter{page}{1}

\setcounter{section}{0}
\setcounter{figure}{0}
\setcounter{table}{0}
\setcounter{equation}{0}

\maketitlesupplementary

\begin{table*}[ht!]
    \centering
    \footnotesize
        \begin{tabular}{lcccc}
            \toprule
            \textbf{Method} & \textbf{ViT Content $\uparrow$} & \textbf{ViT Style $\uparrow$} & \textbf{BLIP-2 Content $\uparrow$} & \textbf{BLIP-2 Style $\uparrow$}\\
            \midrule
            \textbf{ZipLoRA}~\cite{shah2024ziplora} & 23.2\% & 18.0\% & 48.2\% & 46.0\%\\
            \textbf{B-LoRA}~\cite{frenkel2024implicit} & 29.6\% & 19.9\% & 42.0\% & 54.3\%\\
            \textbf{K-LoRA}~\cite{ouyang2025k} & 32.8\% & 20.7\% & 49.4\% & 51.1\%\\
            \textbf{Ours} & 33.7\% & 21.8\% & 49.5\% & 51.4\% \\
            \bottomrule
        \end{tabular}%
        \vspace{-2mm}
    \caption{Comparison of alignment results with additional evaluation metrics. }
    \label{table:more metircs}
    \vspace{-4mm}
\end{table*}

 \begin{figure*}[th!]
	\begin{center}
    \includegraphics[width=0.8\textwidth]{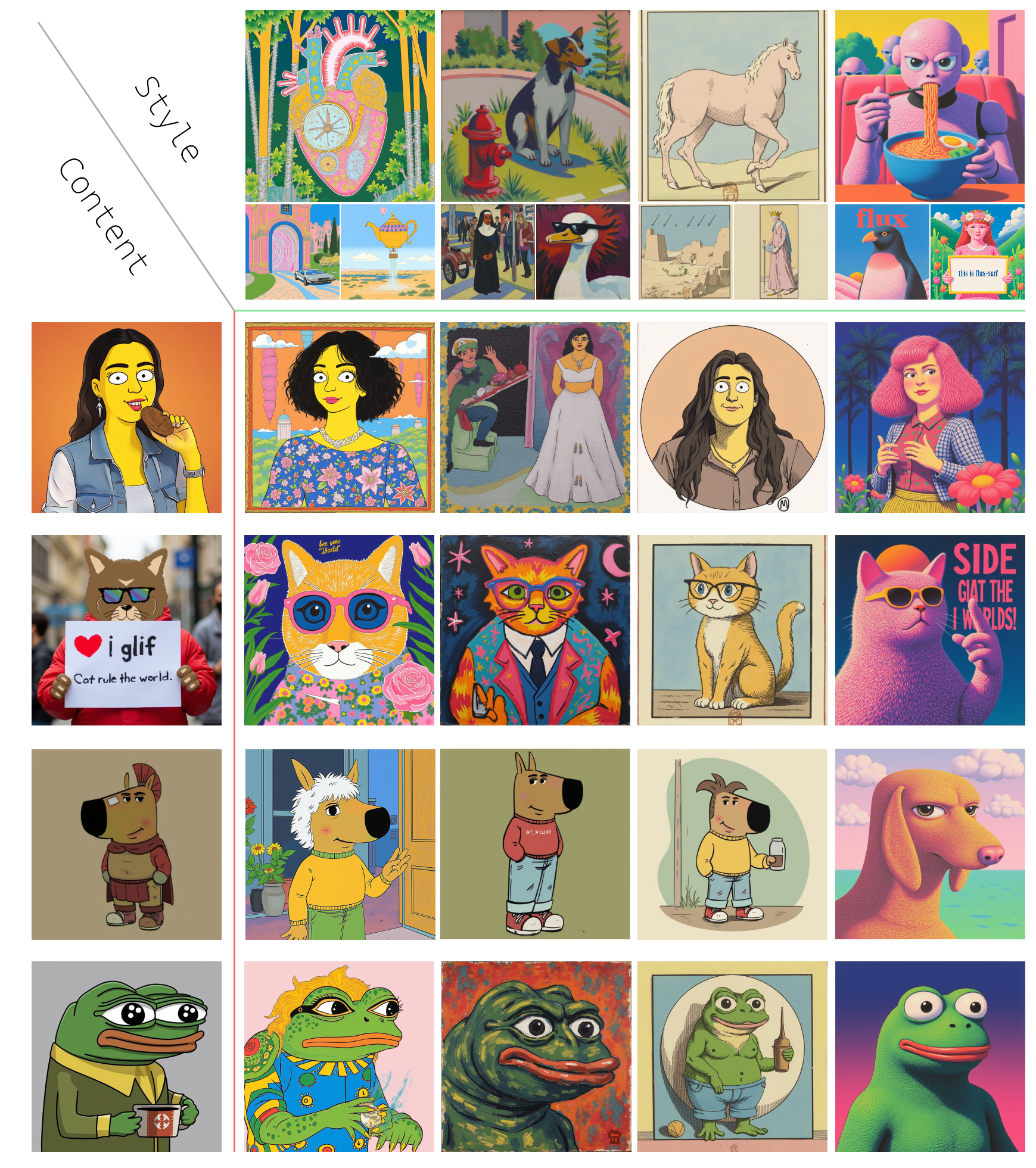}
	\end{center}
	\caption{Additional results generated using FLUX. Each image corresponds to the object label indicated above and the style reference on the left. The results demonstrate the effects of applying different LoRA modules through our proposed method.}
	\label{figs1}
\end{figure*}

\begin{figure*}[th!]
	\begin{center}
    \includegraphics[width=0.8\textwidth]{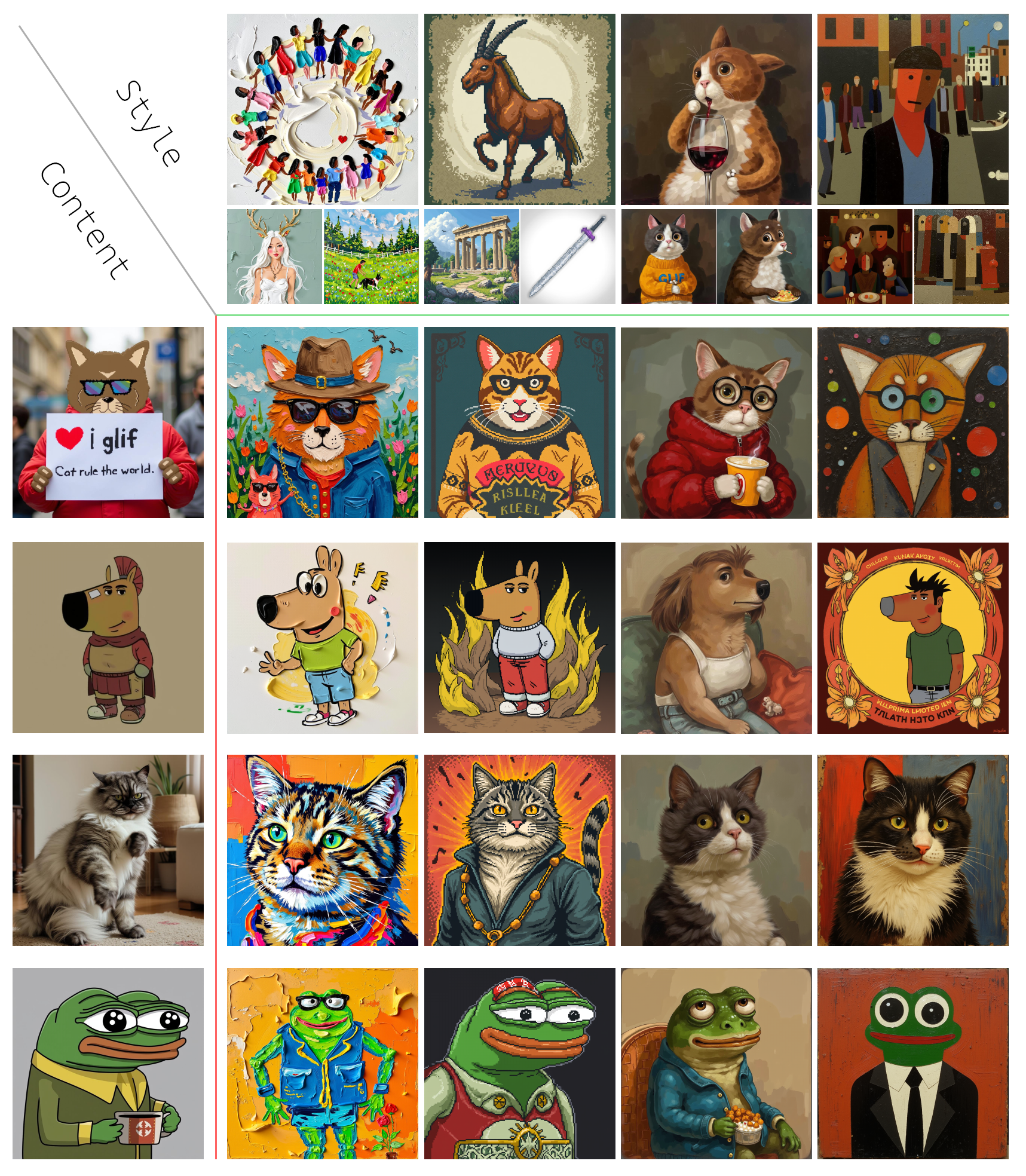}
	\end{center}
	\caption{Additional results generated using FLUX. Each image corresponds to the object label indicated above and the style reference on the left. The results demonstrate the effects of applying different LoRA modules through our proposed method.}
	\label{figs2}
\end{figure*}

\section{Evaluation with Additional Metrics}
\label{Add metircs}
In addition to conventional measures such as Style Similarity, CLIP Score, and DINO Score, we further adopt ViT-based~\cite{dosovitskiyimage} and BLIP-2-based~\cite{li2023blip} content and style scores to provide a more comprehensive evaluation of our proposed method. As shown in Table~\ref{table:more metircs}, our method achieves state-of-the-art results in three out of four metrics. A closer look reveals that different baselines exhibit complementary strengths—ZipLoRA favors BLIP-2 content alignment, while B-LoRA excels in BLIP-2 style consistency. K-LoRA maintains relatively balanced performance across metrics. In contrast, our proposed method not only improves both ViT-based and BLIP-2-based scores, but also demonstrates a better balance between content fidelity and style preservation. We attribute this advantage to the proposed soft, inference-time metric guidance, which adaptively calibrates the generation process to maintain semantic and stylistic coherence without overfitting to a single representation space. These results suggest that our dynamic fusion framework is more robust across heterogeneous evaluation perspectives, highlighting its potential for broader generalization to multimodal generation scenarios.

\section{Additional Experimental Results Based on Flux}
\label{Results on Flux}
As discussed in the main experiments section, to more comprehensively illustrate the superior performance and generalization capability of our proposed method built upon the FLUX framework, we further conduct extensive qualitative evaluations using publicly available, well-trained LoRA (Low-Rank Adaptation) weights shared by the community on HuggingFace. Specifically, we selected a diverse set of LoRA weights corresponding to various object categories and artistic styles to systematically evaluate our model’s ability to integrate and synthesize complex cross-domain representations. The resulting fused images, presented in Figure~\ref{figs1} and Figure~\ref{figs2}, showcase a wide range of combinations where object semantics and stylistic attributes are jointly encoded and rendered through our method.

Our approach incorporates these LoRA weights by disentangling and recombining object- and style-specific latent representations in a manner that performs feature-level selection based on representation perturbation and applying metric-guided refinement during denoising. This enables the model to align and synthesize visual content in a controlled yet flexible fashion. The generated samples exhibit not only strong fidelity to the semantic structure of the target object but also high consistency with the desired style, demonstrating the model's ability to preserve critical attributes from both input domains. Furthermore, the seamless integration of appearance and content substantiates the robustness of our framework in handling varied and unseen combinations, emphasizing its potential applicability in real-world generation tasks that demand stylistic generalization and compositional creativity. Overall, these visual results provide compelling evidence of the effectiveness of our method in producing coherent, high-quality outputs across a broad spectrum of challenging scenarios.

\section{Setup Details for User Study and MLLM Evaluations}
\label{Setup Details}
Following the evaluation protocol of K-LoRA~\cite{ouyang2025k}, we conducted a user study where participants were presented with a \textbf{reference subject image}, a \textbf{reference style image}, and two anonymized outputs—one generated by \textbf{our method} and the other by a randomly selected baseline (ZipLoRA~\cite{shah2024ziplora}, B-LoRA~\cite{frenkel2024implicit}, or K-LoRA~\cite{ouyang2025k}). To mitigate presentation bias, the order of the two outputs was randomized across trials. Participants were asked: \textit{``Which image better reflects the given artistic style while preserving the subject identity?''} We collected a total of 1,290 responses from 43 participants, with each participant evaluating a unique set of 30 trials.

Beyond human evaluation, we further adopt \textbf{GPT-4o} and \textbf{Qwen2.5-VL} as multimodal large language model (MLLM) judges to assess perceptual alignment. For each trial, the prompt included the content image, the style reference, and four anonymized outputs from ZipLoRA, B-LoRA, K-LoRA, and our proposed method(randomized order). The LLMs were instructed to select the image that best balances style fidelity with subject preservation. To ensure robustness, we randomly sampled \textbf{100 subject–style pairs}, and each pair was evaluated in three independent runs. The final score was determined via majority voting across runs, following standard practice in recent MLLM-based evaluation studies. Importantly, while the user study captures subjective human preference, the MLLM-based evaluation provides scalable and reproducible judgments, making the two evaluations complementary and mutually reinforcing.  

%% file: main.bib
@article{zhang2023uncovering,
  title={Uncovering prototypical knowledge for weakly open-vocabulary semantic segmentation},
  author={Zhang, Fei and Zhou, Tianfei and Li, Boyang and He, Hao and Ma, Chaofan and Zhang, Tianjiao and Yao, Jiangchao and Zhang, Ya and Wang, Yanfeng},
  journal={Advances in Neural Information Processing Systems},
  volume={36},
  pages={73652--73665},
  year={2023}
}

@inproceedings{ruiz2023dreambooth,
  title={Dreambooth: Fine tuning text-to-image diffusion models for subject-driven generation},
  author={Ruiz, Nataniel and Li, Yuanzhen and Jampani, Varun and Pritch, Yael and Rubinstein, Michael and Aberman, Kfir},
  booktitle={Proceedings of the IEEE/CVF Conference on Computer Vision and Pattern Recognition},
  pages={22500--22510},
  year={2023}
}

@article{sohn2023styledrop,
  title={Styledrop: Text-to-image synthesis of any style},
  author={Sohn, Kihyuk and Jiang, Lu and Barber, Jarred and Lee, Kimin and Ruiz, Nataniel and Krishnan, Dilip and Chang, Huiwen and Li, Yuanzhen and Essa, Irfan and Rubinstein, Michael and others},
  journal={Advances in Neural Information Processing Systems},
  volume={36},
  pages={66860--66889},
  year={2023}
}

@article{hu2022lora,
  title={Lora: Low-rank adaptation of large language models.},
  author={Hu, Edward J and Shen, Yelong and Wallis, Phillip and Allen-Zhu, Zeyuan and Li, Yuanzhi and Wang, Shean and Wang, Lu and Chen, Weizhu and others},
  journal={International Conference on Learning Representations},
  volume={1},
  number={2},
  pages={3},
  year={2022}
}

@inproceedings{shah2024ziplora,
  title={Ziplora: Any subject in any style by effectively merging loras},
  author={Shah, Viraj and Ruiz, Nataniel and Cole, Forrester and Lu, Erika and Lazebnik, Svetlana and Li, Yuanzhen and Jampani, Varun},
  booktitle={European Conference on Computer Vision},
  pages={422--438},
  year={2024},
  organization={Springer}
}

@inproceedings{frenkel2024implicit,
  title={Implicit style-content separation using b-lora},
  author={Frenkel, Yarden and Vinker, Yael and Shamir, Ariel and Cohen-Or, Daniel},
  booktitle={European Conference on Computer Vision},
  pages={181--198},
  year={2024},
  organization={Springer}
}

@article{ouyang2025k,
  title={K-lora: Unlocking training-free fusion of any subject and style loras},
  author={Ouyang, Ziheng and Li, Zhen and Hou, Qibin},
  journal={arXiv:2502.18461},
  year={2025}
}

@article{ma2023diffusionseg,
  title={Diffusionseg: Adapting diffusion towards unsupervised object discovery},
  author={Ma, Chaofan and Yang, Yuhuan and Ju, Chen and Zhang, Fei and Liu, Jinxiang and Wang, Yu and Zhang, Ya and Wang, Yanfeng},
  journal={arXiv:2303.09813},
  year={2023}
}

@inproceedings{chen2024gentron,
  title={Gentron: Diffusion transformers for image and video generation},
  author={Chen, Shoufa and Xu, Mengmeng and Ren, Jiawei and Cong, Yuren and He, Sen and Xie, Yanping and Sinha, Animesh and Luo, Ping and Xiang, Tao and Perez-Rua, Juan-Manuel},
  booktitle={Proceedings of the IEEE/CVF Conference on Computer Vision and Pattern Recognition},
  pages={6441--6451},
  year={2024}
}

@inproceedings{jiang2024scedit,
  title={Scedit: Efficient and controllable image diffusion generation via skip connection editing},
  author={Jiang, Zeyinzi and Mao, Chaojie and Pan, Yulin and Han, Zhen and Zhang, Jingfeng},
  booktitle={Proceedings of the IEEE/CVF Conference on Computer Vision and Pattern Recognition},
  pages={8995--9004},
  year={2024}
}

@inproceedings{gupta2024photorealistic,
  title={Photorealistic video generation with diffusion models},
  author={Gupta, Agrim and Yu, Lijun and Sohn, Kihyuk and Gu, Xiuye and Hahn, Meera and Li, Fei-Fei and Essa, Irfan and Jiang, Lu and Lezama, Jos{\'e}},
  booktitle={European Conference on Computer Vision},
  pages={393--411},
  year={2024},
  organization={Springer}
}

@article{xing2024survey,
  title={A survey on video diffusion models},
  author={Xing, Zhen and Feng, Qijun and Chen, Haoran and Dai, Qi and Hu, Han and Xu, Hang and Wu, Zuxuan and Jiang, Yu-Gang},
  journal={ACM Computing Surveys},
  volume={57},
  number={2},
  pages={1--42},
  year={2024},
  publisher={ACM New York, NY}
}

@article{gal2022image,
  title={An image is worth one word: Personalizing text-to-image generation using textual inversion},
  author={Gal, Rinon and Alaluf, Yuval and Atzmon, Yuval and Patashnik, Or and Bermano, Amit H and Chechik, Gal and Cohen-Or, Daniel},
  journal={arXiv:2208.01618},
  year={2022}
}

@inproceedings{kumari2023multi,
  title={Multi-concept customization of text-to-image diffusion},
  author={Kumari, Nupur and Zhang, Bingliang and Zhang, Richard and Shechtman, Eli and Zhu, Jun-Yan},
  booktitle={Proceedings of the IEEE/CVF Conference on Computer Vision and Pattern Recognition},
  pages={1931--1941},
  year={2023}
}

@inproceedings{avrahami2023break,
  title={Break-a-scene: Extracting multiple concepts from a single image},
  author={Avrahami, Omri and Aberman, Kfir and Fried, Ohad and Cohen-Or, Daniel and Lischinski, Dani},
  booktitle={SIGGRAPH Asia 2023 Conference Papers},
  pages={1--12},
  year={2023}
}

@inproceedings{shi2024instantbooth,
  title={Instantbooth: Personalized text-to-image generation without test-time finetuning},
  author={Shi, Jing and Xiong, Wei and Lin, Zhe and Jung, Hyun Joon},
  booktitle={Proceedings of the IEEE/CVF conference on computer vision and pattern recognition},
  pages={8543--8552},
  year={2024}
}

@article{xiao2024fastcomposer,
  title={Fastcomposer: Tuning-free multi-subject image generation with localized attention},
  author={Xiao, Guangxuan and Yin, Tianwei and Freeman, William T and Durand, Fr{\'e}do and Han, Song},
  journal={International Journal of Computer Vision},
  pages={1--20},
  year={2024},
  publisher={Springer}
}

@inproceedings{xie2023smartbrush,
  title={Smartbrush: Text and shape guided object inpainting with diffusion model},
  author={Xie, Shaoan and Zhang, Zhifei and Lin, Zhe and Hinz, Tobias and Zhang, Kun},
  booktitle={Proceedings of the IEEE/CVF Conference on Computer Vision and Pattern Recognition},
  pages={22428--22437},
  year={2023}
}

@article{zhang2023lora,
  title={Lora-fa: Memory-efficient low-rank adaptation for large language models fine-tuning},
  author={Zhang, Longteng and Zhang, Lin and Shi, Shaohuai and Chu, Xiaowen and Li, Bo},
  journal={arXiv:2308.03303},
  year={2023}
}

@article{zhou2024lora,
  title={Lora-drop: Efficient lora parameter pruning based on output evaluation},
  author={Zhou, Hongyun and Lu, Xiangyu and Xu, Wang and Zhu, Conghui and Zhao, Tiejun and Yang, Muyun},
  journal={arXiv:2402.07721},
  year={2024}
}

@article{zi2023delta,
  title={Delta-lora: Fine-tuning high-rank parameters with the delta of low-rank matrices},
  author={Zi, Bojia and Qi, Xianbiao and Wang, Lingzhi and Wang, Jianan and Wong, Kam-Fai and Zhang, Lei},
  journal={arXiv:2309.02411},
  year={2023}
}

@article{dong2024continually,
  title={How to Continually Adapt Text-to-Image Diffusion Models for Flexible Customization?},
  author={Dong, Jiahua and Liang, Wenqi and Li, Hongliu and Zhang, Duzhen and Cao, Meng and Ding, Henghui and Khan, Salman H and Shahbaz Khan, Fahad},
  journal={Advances in Neural Information Processing Systems},
  volume={37},
  pages={130057--130083},
  year={2024}
}

@article{gu2023mix,
  title={Mix-of-show: Decentralized low-rank adaptation for multi-concept customization of diffusion models},
  author={Gu, Yuchao and Wang, Xintao and Wu, Jay Zhangjie and Shi, Yujun and Chen, Yunpeng and Fan, Zihan and Xiao, Wuyou and Zhao, Rui and Chang, Shuning and Wu, Weijia and others},
  journal={Advances in Neural Information Processing Systems},
  volume={36},
  pages={15890--15902},
  year={2023}
}

@article{jiang2024mc,
  title={MC2: Multi-concept Guidance for Customized Multi-concept Generation},
  author={Jiang, Jiaxiu and Zhang, Yabo and Feng, Kailai and Wu, Xiaohe and Li, Wenbo and Pei, Renjing and Li, Fan and Zuo, Wangmeng},
  journal={arXiv:2404.05268},
  year={2024}
}

@article{xing2024csgo,
  title={Csgo: Content-style composition in text-to-image generation},
  author={Xing, Peng and Wang, Haofan and Sun, Yanpeng and Wang, Qixun and Bai, Xu and Ai, Hao and Huang, Renyuan and Li, Zechao},
  journal={arXiv:2408.16766},
  year={2024}
}

@article{wu2024mixture,
  title={Mixture-of-subspaces in low-rank adaptation},
  author={Wu, Taiqiang and Wang, Jiahao and Zhao, Zhe and Wong, Ngai},
  journal={arXiv:2406.11909},
  year={2024}
}

@inproceedings{saharia2022palette,
  title={Palette: Image-to-image diffusion models},
  author={Saharia, Chitwan and Chan, William and Chang, Huiwen and Lee, Chris and Ho, Jonathan and Salimans, Tim and Fleet, David and Norouzi, Mohammad},
  booktitle={ACM SIGGRAPH 2022 conference proceedings},
  year={2022}
}

@article{kazerouni2022diffusion,
  title={Diffusion models for medical image analysis: A comprehensive survey},
  author={Kazerouni, Amirhossein and Aghdam, Ehsan Khodapanah and Heidari, Moein and Azad, Reza and Fayyaz, Mohsen and Hacihaliloglu, Ilker and Merhof, Dorit},
  journal={arXiv:2211.07804},
  year={2022}
}

@inproceedings{chen2023diffusiondet,
  title={Diffusiondet: Diffusion model for object detection},
  author={Chen, Shoufa and Sun, Peize and Song, Yibing and Luo, Ping},
  booktitle={Proceedings of the IEEE/CVF International Conference on Computer Vision},
  year={2023}
}

@article{amit2021segdiff,
  title={Segdiff: Image segmentation with diffusion probabilistic models},
  author={Amit, Tomer and Shaharbany, Tal and Nachmani, Eliya and Wolf, Lior},
  journal={arXiv:2112.00390},
  year={2021}
}

@article{song2020score,
  title={Score-based generative modeling through stochastic differential equations},
  author={Song, Yang and Sohl-Dickstein, Jascha and Kingma, Diederik P and Kumar, Abhishek and Ermon, Stefano and Poole, Ben},
  journal={arXiv:2011.13456},
  year={2020}
}

@article{kaltenbach2020incorporating,
  title={Incorporating physical constraints in a deep probabilistic machine learning framework for coarse-graining dynamical systems},
  author={Kaltenbach, Sebastian and Koutsourelakis, Phaedon-Stelios},
  journal={Journal of Computational Physics},
  volume={419},
  pages={109673},
  year={2020},
  publisher={Elsevier}
}

@article{zhang2022dino,
  title={Dino: Detr with improved denoising anchor boxes for end-to-end object detection},
  author={Zhang, Hao and Li, Feng and Liu, Shilong and Zhang, Lei and Su, Hang and Zhu, Jun and Ni, Lionel M and Shum, Heung-Yeung},
  journal={arXiv:2203.03605},
  year={2022}
}

@inproceedings{radford2021learning,
  title={Learning transferable visual models from natural language supervision},
  author={Radford, Alec and Kim, Jong Wook and Hallacy, Chris and Ramesh, Aditya and Goh, Gabriel and Agarwal, Sandhini and Sastry, Girish and Askell, Amanda and Mishkin, Pamela and Clark, Jack and others},
  booktitle={International Conference on Machine Learning},
  pages={8748--8763},
  year={2021},
  organization={PmLR}
}

@inproceedings{caron2021emerging,
  title={Emerging properties in self-supervised vision transformers},
  author={Caron, Mathilde and Touvron, Hugo and Misra, Ishan and J{\'e}gou, Herv{\'e} and Mairal, Julien and Bojanowski, Piotr and Joulin, Armand},
  booktitle={Proceedings of the IEEE/CVF International Conference on Computer Vision},
  pages={9650--9660},
  year={2021}
}

@article{he2025cameractrl,
  title={CameraCtrl II: Dynamic Scene Exploration via Camera-controlled Video Diffusion Models},
  author={He, Hao and Yang, Ceyuan and Lin, Shanchuan and Xu, Yinghao and Wei, Meng and Gui, Liangke and Zhao, Qi and Wetzstein, Gordon and Jiang, Lu and Li, Hongsheng},
  journal={arXiv:2503.10592},
  year={2025}
}

@article{he2024cameractrl,
  title={Cameractrl: Enabling camera control for text-to-video generation},
  author={He, Hao and Xu, Yinghao and Guo, Yuwei and Wetzstein, Gordon and Dai, Bo and Li, Hongsheng and Yang, Ceyuan},
  journal={arXiv:2404.02101},
  year={2024}
}

@article{cao2024teaching,
  title={Teaching video diffusion model with latent physical phenomenon knowledge},
  author={Cao, Qinglong and Wang, Ding and Li, Xirui and Chen, Yuntian and Ma, Chao and Yang, Xiaokang},
  journal={arXiv:2411.11343},
  year={2024}
}

@article{ho2022video,
  title={Video diffusion models},
  author={Ho, Jonathan and Salimans, Tim and Gritsenko, Alexey and Chan, William and Norouzi, Mohammad and Fleet, David J},
  journal={Advances in Neural Information Processing Systems},
  volume={35},
  pages={8633--8646},
  year={2022}
}

@inproceedings{zheng2023layoutdiffusion,
  title={Layoutdiffusion: Controllable diffusion model for layout-to-image generation},
  author={Zheng, Guangcong and Zhou, Xianpan and Li, Xuewei and Qi, Zhongang and Shan, Ying and Li, Xi},
  booktitle={Proceedings of the IEEE/CVF Conference on Computer Vision and Pattern Recognition},
  pages={22490--22499},
  year={2023}
}

@misc{openai2024gpt4o,
  title        = {Introducing GPT-4o: OpenAI's New Flagship Multimodal Model},
  author       = {{OpenAI} and Microsoft},
  howpublished = {Azure Blog},
  url          = {https://azure.microsoft.com/en-us/blog/
                  introducing-gpt-4o-openais-new
                  /-flagship-multimodal-model-now-in-preview-on-azure/},
  month        = may,
  year         = {2024},
}

@article{Qwen2.5-VL,
  title={Qwen2.5-VL Technical Report},
  author={Bai, Shuai and Chen, Keqin and Liu, Xuejing and Wang, Jialin and Ge, Wenbin and Song, Sibo and Dang, Kai and Wang, Peng and Wang, Shijie and Tang, Jun and Zhong, Humen and Zhu, Yuanzhi and Yang, Mingkun and Li, Zhaohai and Wan, Jianqiang and Wang, Pengfei and Ding, Wei and Fu, Zheren and Xu, Yiheng and Ye, Jiabo and Zhang, Xi and Xie, Tianbao and Cheng, Zesen and Zhang, Hang and Yang, Zhibo and Xu, Haiyang and Lin, Junyang},
  journal={arXiv:2502.13923},
  year={2025}
}

@inproceedings{dosovitskiyimage,
  title={An Image is Worth 16x16 Words: Transformers for Image Recognition at Scale},
  author={Dosovitskiy, Alexey and Beyer, Lucas and Kolesnikov, Alexander and Weissenborn, Dirk and Zhai, Xiaohua and Unterthiner, Thomas and Dehghani, Mostafa and Minderer, Matthias and Heigold, Georg and Gelly, Sylvain and others},
  booktitle={International Conference on Learning Representations},
  year={2021}
}

@inproceedings{li2023blip,
  title={Blip-2: Bootstrapping language-image pre-training with frozen image encoders and large language models},
  author={Li, Junnan and Li, Dongxu and Savarese, Silvio and Hoi, Steven},
  booktitle={International Conference on Machine Learning},
  pages={19730--19742},
  year={2023},
  organization={PMLR}
}

@inproceedings{xu2025personalized,
  title={Personalized image generation with large multimodal models},
  author={Xu, Yiyan and Wang, Wenjie and Zhang, Yang and Tang, Biao and Yan, Peng and Feng, Fuli and He, Xiangnan},
  booktitle={Proceedings of the ACM on Web Conference 2025},
  pages={264--274},
  year={2025}
}

@inproceedings{dong2025insight,
  title={Insight-v: Exploring long-chain visual reasoning with multimodal large language models},
  author={Dong, Yuhao and Liu, Zuyan and Sun, Hai-Long and Yang, Jingkang and Hu, Winston and Rao, Yongming and Liu, Ziwei},
  booktitle={Proceedings of the Computer Vision and Pattern Recognition Conference},
  pages={9062--9072},
  year={2025}
}

@article{lu2025ovis2,
  title={Ovis2. 5 technical report},
  author={Lu, Shiyin and Li, Yang and Xia, Yu and Hu, Yuwei and Zhao, Shanshan and Ma, Yanqing and Wei, Zhichao and Li, Yinglun and Duan, Lunhao and Zhao, Jianshan and others},
  journal={arXiv:2508.11737},
  year={2025}
}
